\documentclass[preprint,12pt]{elsarticle}

\usepackage{amssymb}
\usepackage{amsmath}
\usepackage{graphicx}%
\usepackage{multirow}%
\usepackage{amsmath,amssymb,amsfonts}%
\usepackage{amsthm}%
\usepackage{mathrsfs}%
\usepackage[title]{appendix}%
\usepackage{xcolor}%
\usepackage{textcomp}%
\usepackage{manyfoot}%
\usepackage{booktabs}%
\usepackage{algorithm}%
\usepackage{algorithmicx}%
\usepackage{algpseudocode}%
\usepackage{listings}%
\usepackage{pifont}%
\usepackage{hyperref}%

\journal{}

\begin{document}

\begin{frontmatter}

\title{Efficient Point Cloud Classification via Offline Distillation Framework and Negative-Weight Self-Distillation Technique}

\author[label1]{Qiang Zheng}
\author[label1]{Chao Zhang}
\author[label1]{Jian Sun \corref{cor1}}

\begin{abstract}
The rapid advancement in point cloud processing technologies has significantly increased the demand for efficient and compact models that achieve high-accuracy classification. Knowledge distillation has emerged as a potent model compression technique. However, traditional KD often requires extensive computational resources for forward inference of large teacher models, thereby reducing training efficiency for student models and increasing resource demands. To address these challenges, we introduce an innovative offline recording strategy that avoids the simultaneous loading of both teacher and student models, thereby reducing hardware demands. This approach feeds a multitude of augmented samples into the teacher model, recording both the data augmentation parameters and the corresponding logit outputs. By applying shape-level augmentation operations such as random scaling and translation, while excluding point-level operations like random jittering, the size of the records is significantly reduced. Additionally, to mitigate the issue of small student model over-imitating the teacher model's outputs and converging to suboptimal solutions, we incorporate a negative-weight self-distillation strategy. Experimental results demonstrate that the proposed distillation strategy enables the student model to achieve performance comparable to state-of-the-art models while maintaining lower parameter count. This approach strikes an optimal balance between performance and complexity. This study highlights the potential of our method to optimize knowledge distillation for point cloud classification tasks, particularly in resource-constrained environments, providing a novel solution for efficient point cloud analysis.
\end{abstract}



\end{frontmatter}

\section{Introduction}

Knowledge distillation (KD)~\cite{2015KD} is an effective model compression method. However, applying KD to point cloud classification models faces numerous challenges. Traditional KD methods often rely on high-capacity teacher models, which consume significant computational resources during forward inference, restricting the efficiency of student model training. Additionally, these methods require frequent loading of teacher models for inference across various application scenarios, making them ineffective in resource-constrained environments. Moreover, student models may converge too quickly to suboptimal solutions while imitating teacher models, limiting the effectiveness of distillation in practical scenarios.

This study proposes an offline distillation framework. Initially, the pre-trained teacher model is loaded separately to generate a diverse set of point cloud samples through various data augmentation operations. These samples undergo logit output inference by the teacher model. During this process, the augmentation parameters and corresponding logit outputs are recorded offline for subsequent student model training. The study employs holistic augmentation methods such as random scaling and translation to ensure that all points in one sample share the same augmentation parameters. The proposed offline distillation framework achieves asynchronous loading of teacher and student models, reducing resource requirements.

The offline record strategy in this framework enhances the flexibility of other knowledge distillation schemes. Pre-storing a large number of records corresponding to augmented samples creates a rich repository of knowledge from the teacher. These records can be used for knowledge distillation in this study and other application scenarios, improving data reusability, avoiding repeated loading of the teacher model, and reducing computational resource consumption.

This study also introduces a negative-weight self-distillation technique to improve the generalization ability of the student model. In traditional self-distillation~\cite{2022self_distil}, the student model learns by imitating its outputs from previous iterations. When the external teacher model has a strong guiding effect, self-distillation may exacerbate insufficient student model training. Negative-weight self-distillation encourages the student model to produce different logit outputs for input samples in consecutive iterations by introducing a negative coefficient in the self-distillation loss term. This strategy forces the model to explore a broader feature space, learning more robust and diverse feature representations. Negative-weight self-distillation provides a new regularization mechanism, balancing the model's training process and preventing a decrease in adaptability to new situations in the pursuit of precise imitation.

\section{Related works}	
\subsection{Point Cloud Analysis}
Point cloud analysis constitutes a fundamental technique in 3D data processing, with research endeavors delineated into three primary categories based on their treatment of point cloud data: projection-based methods, voxel-based methods, and point-based methods. 

Point-based methods directly process the original point cloud data without undergoing data transformation or mapping. The primary advantage of these methods lies in their ability to fully exploit the inherent structure of the point cloud data, thereby avoiding potential information loss during the conversion process. Point-based methods are further classified into Multi-Layer Perceptron (MLP)-based, Convolutional Neural Network (CNN)-based, Graph Neural Network (GNN)-based, and Transformer-based approaches. The MLP-based method~\cite{PointNet2017, PointNetplus2017} utilizes shared MLPs to transform point features and employs pooling operations to aggregate features. Although this strategy overlooks the inherent connections among points, MLP-based methods demonstrate relatively high computational efficiency. The CNN-based method~\cite{2018PointCNN, 2021Deep} learns convolutional kernels for local areas and extracts features through convolutional operations, proficient in capturing local patterns within the point cloud. The GNN-based method~\cite{DGCNN2019, Adaptive2021} treats the point cloud as graph data, extracting features from nodes and edges and aggregating node information in a high-dimensional feature space, thereby being apt for complex topological structure analysis. Finally, the Transformer-based method~\cite{point-trans2021, point-transV2-2022} perceives the point cloud as an unordered set of points, dynamically generating weights for each point using attention mechanisms, offering high flexibility and scalability. In the realm of point cloud analysis, while prior works have made significant strides, they often prioritize either efficiency or accuracy. Our study, however, seeks to strike a balance between these two critical aspects, aiming for a model that delivers both high performance and computational efficiency.

\subsection{Knowledge Distillation}
In the realm of knowledge distillation, seminal works have significantly advanced both logit and feature distillation techniques. KD~\cite{2015KD} introduces knowledge distillation, inspiring subsequent research. Yang et al. proposed SRRL~\cite{2021SSRL}, optimizing penultimate layer features of student networks. Zhao et al. introduce DKD~\cite{2022DKD}, offering independent optimization of target and non-target class knowledge. KDExplainer~\cite{2021KDExplainer} by Xue et al. explores attention mechanisms, providing deeper insights into the distillation process. Cui et al. integrate self-supervision tasks with knowledge distillation~\cite{2020SSKD}, enhancing knowledge transfer in various scenarios. FN~\cite{2020FN} proposed by Xu et al. addressed label noise using feature normalization. Liu et al. propose SimKD~\cite{2022SimKD}, reusing the teacher's classifier for student inference. Zhou et al.~\cite{2021WSLD} addresse the bias-variance tradeoff in knowledge distillation using weighted soft labels. Zagoruyko and Komodakis~\cite{2023NKD} focuse on attention transfer for knowledge transfer in convolutional neural networks. Moreover, advancements in feature distillation have been made. Romero et al. introduce FitNets~\cite{2015FitNets}, utilizing intermediate representations to guide training of thinner student networks. Park et al. propose RKD~\cite{2019RKD}, transferring mutual relations between data examples. Tung and Mori introduce SPKD~\cite{2019SP}, preserving pairwise similarity of input activations. Zagoruyko and Komodakis~\cite{2017AT} explore attention transfer for improved CNN performance. Chen et al.~\cite{2021reviewKD} addresse semantic mismatch with a knowledge review framework emphasizing cross-layer information flow. Chen et al. propose CLDSC~\cite{2021Cross}, alleviating semantic mismatch and improving generalization. Distinguished from existing knowledge distillation methods, our study introduces a negative-weight self-distillation technique. This innovative approach is tailored to prevent the student model from converging prematurely on suboptimal solutions, thereby fostering a more robust and diverse feature learning process.

\section{Methodology}

\subsection{Offline Distillation Framework}
\begin{figure}[htbp]
    \centering
    \includegraphics[width=0.99\linewidth]{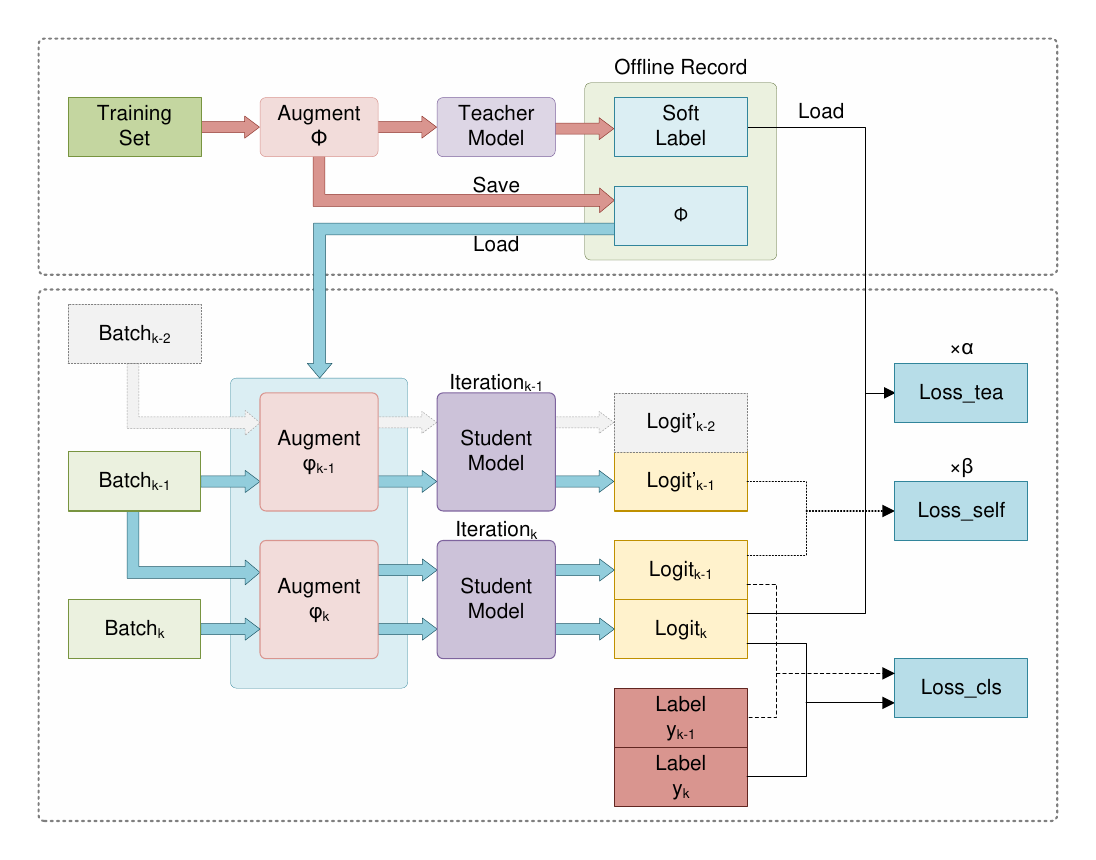}
    \caption{The figure depicts the two-stage offline distillation framework, beginning with a pre-trained teacher model that infers input samples and captures soft labels and data augmentation parameters. The subsequent phase trains a student model utilizing the offline record for teacher-student distillation, enhanced by the introduction of negative weight self-distillation. The overall architecture encompasses three types of loss functions: classification, teacher-student distillation, and self-distillation.}
    \label{fig-PointDistil-framework}
\end{figure}

In contrast to traditional knowledge distillation methods, which often involve synchronous loading and inference with a teacher model, our study introduces an offline distillation framework, illustrated in Fig.~\ref{fig-PointDistil-framework}. This framework addresses the high computational and storage demands associated with real-time inference of large teacher models. The general process involves generating offline records in the first phase, which are then utilized for training the student model in a subsequent phase. This separation enables the student model to benefit from the teacher's knowledge without requiring the teacher model's presence during training, thus reducing overall resource consumption.

Creating offline records involves multiple epochs of inference with the teacher model. Unlike traditional methods that record soft labels directly from the original, unaltered training set, our method ensures that each epoch's results correspond to a complete set of records. During this phase, data augmentation techniques such as random scaling and translation are applied without shuffling the training set, ensuring that each record is aligned with the original sample order. This approach captures the nuanced changes introduced by data augmentation, which are essential for the student model to learn from varied data representations. By recording the augmented samples alongside their corresponding soft labels, we establish a robust linkage that is critical for effective knowledge transfer.

The utilization of offline records in the student model's training follows a process of selective sampling from the pre-generated records. For each training epoch, a set of records, including the augmentation parameters and the corresponding soft labels, is randomly selected. If the training set undergoes any shuffling operations, the selected records are shuffled in the same manner to ensure synchronicity between the samples, augmentation parameters, and soft labels. This synchronization preserves the one-to-one correspondence required for accurate training.

The offline distillation framework and associated offline records offer several distinct advantages. Firstly, they allow for a more flexible and efficient allocation of computational resources, as the teacher model's inference need not be concurrent with the student model's training. Secondly, by generating an excess of augmented samples, the student model is exposed to a wider array of data representations, enhancing its generalization capabilities. Lastly, the offline records can be reused across different training sessions and with various student models, increasing the scalability and adaptability of the knowledge distillation process.

\subsection{Negative-Weight Self-Distillation}

The traditional self-distillation approach, despite its benefits, presents unique challenges when applied to smaller models. These models, with their limited parameters and straightforward architecture, may lack sufficient internal knowledge for effective distillation. This limitation can hinder performance improvement. Additionally, their simple structure and lack of diversity in components might result in only minor learning gains during the self-distillation process. Smaller models, due to their limited representational power, might quickly converge on suboptimal solutions rather than finding the global optimum. In such cases, traditional self-distillation could inadvertently reinforce this tendency by consistently replicating the model's existing knowledge without the necessary variety to explore better solutions.

To address these issues, this study introduces negative-weight self-distillation into the student model's training process, illustrated in Fig.~\ref{fig-PointDistil-framework}. Unlike conventional self-distillation, this method assigns a negative weight to the self-distillation loss. In our approach, the student model receives two sets of data within each training batch: the current batch data \(\mathcal{B}_{k}\) and data from the preceding batch \(\mathcal{B}_{k-1}\). Both datasets are augmented based on the parameters saved by the teacher model for the current batch samples \(\mathcal{B}_{k}\), generating diverse sample representations. The student model computes the classification loss for the augmented \(\mathcal{B}_{k}\) and \(\mathcal{B}_{k-1}\), while also evaluating the KL divergence between the outputs of \(\mathcal{B}_{k-1}\) across two consecutive batches. This KL divergence measures the difference between the two probability distributions and indicates how consistently the model represents the same batch samples across successive batches.

The use of negative-weight self-distillation in our point cloud classification experiment has several advantages. Firstly, it acts as a form of regularization, introducing additional constraints in the loss function that encourage the model to maintain a certain level of dissimilarity in the representation of identical samples across consecutive batches. This helps prevent the model parameters from converging too closely to a local optimum, instead guiding the model to explore a broader parameter space to find a better optimum. By incorporating a negative-weight self-distillation loss, the model is compelled to generate distinct outputs for the same data in consecutive iterations under the influence of the teacher model's soft labels, thereby enhancing its adaptability to new data by acquiring diverse data representations.

In summary, our research introduces a classification model that integrates teacher-student distillation and negative-weight self-distillation, incorporating three distinct components in the loss function. The expressions for these components and the overall loss are specified as follows:
\begin{equation}
\mathcal{L}_{CE} = \frac{1}{n} \mathcal{H}([\mathbf{p}_{i,s}^{pre}, \mathbf{p}_{i,s}^{cur}], [\mathbf{y}_{i}^{pre}, \mathbf{y}_{i}^{cur}])
\label{eq-loss-cls}
\end{equation}

\begin{equation}
\mathcal{L}_{dist}^{(tea)} = \frac{1}{n} \sum_{i} {T_{tea}^2} D_{\text{KL}}(\mathbf{p}^{cur}_{i,s} || \mathbf{p}^{cur}_{i,t})
\end{equation}

\begin{equation}
\mathcal{L}_{dist}^{(self)} = \frac{1}{n} \sum_{i} {T_{self}^2} D_{\text{KL}}(\mathbf{p}^{pre}_{i,s} || \mathbf{p}^{'pre}_{i,s})
\end{equation}

\begin{equation}
\mathcal{L}_{total} = \mathcal{L}_{CE} + \alpha \mathcal{L}_{dist}^{(tea)} + \beta \mathcal{L}_{dist}^{(self)}
\label{eq-loss-total}
\end{equation}

Here, \(\mathcal{L}_{CE}\) represents the cross-entropy loss for classification, \(\mathcal{L}_{dist}^{(tea)}\) is the teacher-student distillation loss, and \(\mathcal{L}_{dist}^{(self)}\) is the self-distillation loss with a negative weight. The variables \(T_{tea}^2\) and \(T_{self}^2\) scale the distillation losses, and the weight coefficients \(\alpha\) and \(\beta\) are set such that \(\alpha > 0\) and \(\beta < 0\). The terms ``cur'' and ``pre'' denote the data corresponding to the current batch and the previous batch, respectively, with ``[ ]'' indicating concatenation. The symbols ``t'' and ``s'' represent the outputs of the teacher and student models. Additionally, \(\mathbf{y}_{i}^{cur}\) and \(\mathbf{y}_{i}^{pre}\), \(\mathbf{p}_{i,s}^{cur}\), and \(\mathbf{p}_{i,s}^{pre}\) denote the ground truths and logits of the \(i\)-th sample in the current and previous batches, respectively. \(\mathbf{p}^{cur}_{i,t}\) indicates the soft labels obtained from the teacher model for the current batch data, and \(\mathbf{p}^{'pre}_{i,s}\) signifies the logits produced for the previous batch data during the training of that previous batch.

\subsection{Network Configuration}
In this study, the teacher network is built upon the PointViG model~\cite{2024PointViG}. A condensed student model denoted as \textbf{PointViG}-\textbf{Distil}lation (PointViG-Distil) is derived by reducing the number of layers in the PointViG~\cite{2024PointViG} framework, with the student model containing only a single graph convolutional module. Additionally, recognizing that the teacher model progressively expands its perceptual field through a hierarchical encoder, the student model adjusts the neighborhood size associated with local embedding and graph convolution. The principal differences between the student and teacher models are outlined in Table \ref{tab-compare-tea-stu}.

\begin{table}[ht]
    \centering
    \small
    \setlength{\tabcolsep}{5.5mm}
    \begin{tabular*}{\textwidth}{@{\extracolsep{\fill}}l|cc}
        \toprule[1pt]
        Encoder Param.                  & PointViG~\cite{2024PointViG}  & PointViG-Distil \\
        \midrule[0.3pt]
        Neigh. Size (Local Embed.)      & 24                        & 48        \\
        Num. Stage                      & 3                         & 1         \\
        Num. Block                      & [1, 1, 1]                 & [1,]      \\
        Channels                        & [64, 128, 256]            & [128,]    \\
        Neigh. Size (Graph Conv.)       & [8, 8, 8]                 & [32,]     \\
        \bottomrule[1pt]
    \end{tabular*}
    \vspace{5pt}
    \caption{Comparison of Encoder Hyperparameters between Teacher and Student Models.}
    \label{tab-compare-tea-stu}
\end{table}

During the training process of the student model, the iteration is set for 100 epochs with a batch size of 32. The optimization process employs the Adam optimizer, with the learning rate initialized at $1.0 \times 10^{-3}$ and decaying to $1.0 \times 10^{-5}$. To ensure robust guidance from the teacher model while preventing the negative-weight self-distillation from hindering the student model's convergence, we set a larger positive value for \(\alpha\) and assigned \(\beta\) a smaller absolute value. Specifically, in the loss function outlined in Eq. \ref{eq-loss-total}, \(\alpha\) is set to 2.0, \(\beta\) is set to \(-0.01\), and the temperatures \(T_{\text{tea}}\) and \(T_{\text{self}}\) are both set to 3.0.

\section{Experiments}

\subsection{ModelNet40 Classification}
The ModelNet40 dataset is used as a benchmark to evaluate our model's performance in point cloud classification. ModelNet40 comprises 40 categories of objects with a total of 12311 unique 3D models. Following PointNet~\cite{PointNet2017}, each model in the dataset is represented as a point cloud with 1024 points. 

We present a comparative analysis of the PointViG~\cite{2024PointViG} model and its distilled variant, PointViG-Distil, on the ModelNet40 classification task, as detailed in Table~\ref{tab-cls}. The PointViG~\cite{2024PointViG} model, serving as our teacher model, demonstrated an accuracy of 94.3\%. The PointViG-Distil model, which incorporates our proposed knowledge distillation techniques, achieved an accuracy of 94.1\%. This result indicates that the distilled model retains a high level of performance while having significantly fewer parameters compared to the original PointViG~\cite{2024PointViG} model. This performance comparison is particularly noteworthy as it highlights the effectiveness of our distillation approach in compressing the model without a substantial loss in accuracy. The close performance between PointViG~\cite{2024PointViG} and PointViG-Distil, despite the latter's reduced parameter count, underscores the success of our method in efficiently transferring knowledge from a larger model to a more compact one.

\begin{table}[ht]
    \centering
    \setlength{\tabcolsep}{5.0mm}
    \begin{tabular*}{\textwidth}{l|cccc}
        \toprule[1pt]
        \textbf{Method} & Input & Num. & mAcc(\%) & OA(\%) \\
        \midrule[0.3pt]        
        PointNet~\cite{PointNet2017}					& xyz           & 1k    & 86.0  & 89.2 \\
        PointNet++(MSG)~\cite{PointNetplus2017}			& xyz, nor      & 5k    & -     & 91.9 \\
        PointCNN~\cite{PointCNN2018}					& xyz           & 1k    & 88.1  & 92.2 \\
        DGCNN~\cite{DGCNN2019}					        & xyz           & 1k    & 90.2  & 92.9 \\
        RS-CNN~\cite{RSCNN2019} w/o vot.                & xyz           & 1k    & -     & 92.9 \\
        KPConv~\cite{KPConv2019}					    & xyz           & 6.8k  & -     & 92.9 \\
        PointNext~\cite{PointNext2022}                  & xyz           & 1k    & 90.8  & 93.2 \\
        AdaptConv~\cite{Adaptive2021}			        & xyz           & 1k    & 90.7  & 93.4 \\
        PointMixer~\cite{pointmixer2021}                & xyz           & 1k    & 91.4  & 93.6 \\
        PT~\cite{point-trans2021}                       & xyz           & 1k    & 90.6  & 93.7 \\
        CurveNet~\cite{CurveNet2021}                    & xyz           & 1k    & -     & 93.8 \\
        PointMLP~\cite{PointMLP2022}                    & xyz           & 1k    & 90.9  & 94.1 \\
        \midrule[0.3pt]
        PointViG~\cite{2024PointViG}                    & xyz           & 1k    & 91.2  & 94.3 \\
        PointViG-Distil (Ours)                          & xyz           & 1k    & 91.0  & 94.1 \\
        \bottomrule[1pt]
    \end{tabular*}
    \vspace{5pt}
    \caption{Results for the ModelNet40 classification task.}
    \label{tab-cls}
\end{table}

\subsection{Complexity Analysis}
Tab.~\ref{tab-complexity} presents a complexity analysis of PointViG~\cite{2024PointViG}. Compared to its teacher model, PointViG-Distil does not show a reduction in time complexity. This is because PointViG~\cite{2024PointViG} is already highly computationally efficient compared to other state-of-the-art models. In the case of PointViG-Distil, the encoder is compressed to a single graph convolution block, and measures have been taken to enhance the perceptual field by increasing the graph neighborhood size, which inevitably increases computational demand. Despite this, PointViG-Distil achieves a significant reduction in parameter count compared to PointViG~\cite{2024PointViG}.
The comparative analysis shows that PointViG-Distil, while maintaining competitive accuracy, optimizes model complexity, which is particularly beneficial for applications with constrained computational budgets.

\begin{table}[ht]
    \centering
    \newcommand{\tabincell}[2]{\begin{tabular}{@{}#1@{}}#2\end{tabular}}
    \setlength{\tabcolsep}{5.5mm}
    \begin{tabular*}{\textwidth}{l|cccc}
        \toprule[1pt]
        \textbf{Method}     & \tabincell{c}{Params.\\(M)}   & \tabincell{c}{FLOPs\\(G)}
                            & \tabincell{c}{mAcc\\(\%)}     & \tabincell{c}{OA\\(\%)} \\  
        \midrule[0.3pt]
        PointNet~\cite{PointNet2017}					& 3.5           & 0.9   & 86.2  & 89.2 \\
        PointNet++(MSG)~\cite{PointNetplus2017}			& 1.7           & 4.1   & -     & 91.9 \\
        PointCNN~\cite{PointCNN2018}					& 0.6           & -     & 88.1  & 92.2 \\
        DGCNN~\cite{DGCNN2019}					        & 1.8           & 4.8   & 90.2  & 92.9 \\
        DeepGCN~\cite{DeepGCNs2021}                     & 2.2           & 3.9   & 90.9  & 93.6 \\
        PointNext-S~\cite{PointNext2022}                & 4.5           & 6.5   & 90.9  & 93.7 \\
        PointMLP~\cite{PointMLP2022}                    & 13.2          & 31.3  & 90.9  & 94.1 \\
        PointWavelet-L~\cite{PointWavelet2023}          & 58.4          & 39.2  & 91.1  & 94.3\\
        \midrule[0.3pt]
        PointViG~\cite{2024PointViG}                    & 1.5           & 0.6   & 91.2  & 94.3 \\
        PointViG-Distil (Ours)                          & 0.4           & 0.6   & 91.0  & 94.1 \\       
        \bottomrule[1pt]
    \end{tabular*}
    \vspace{5pt}
    \caption{Results of the complexity analysis on ModelNet40 classification task (M: $10^{6}$, G: $10^{9}$). Due to the enlargement of neighborhood sizes, the FLOPs of PointViG-Distil have not decreased compared to the teacher model; however, the number of parameters has been significantly reduced.}
    \label{tab-complexity}
\end{table}

\subsection{Ablation Experiments on Framework Design}
The ablation experiment was designed to elucidate the role of the distillation framework, with results documented in Tab.~\ref{tab-ablation}. This table presents the performance of four configurations: Model-1 (No Distill.), Model-2 (Tea. Distil.), Model-3 (Self. Distil.), and PointViG-Distil (Tea. \& Self. Distil.). These models correspond to scenarios with no distillation, teacher-student distillation only, negative-weight self-distillation only, and the integration of both distillation methods, respectively. The accuracy rates for these models are recorded as 92.6\%, 93.9\%, 92.6\%, and 94.1\%, respectively.

\begin{table}[ht]
    \centering
    \newcommand{\tabincell}[2]{\begin{tabular}{@{}#1@{}}#2\end{tabular}}
    \small
    \setlength{\tabcolsep}{3.0mm}
    \begin{tabular*}{\textwidth}{c|cccc}
        \toprule[1pt]
        Model   & \tabincell{c}{Model-1\\(No Distill.)}     & \tabincell{c}{Model-2\\(Tea. Distil.)}  	
                & \tabincell{c}{Model-3\\(Self. Distil.)}  	& \tabincell{c}{PointViG-Distil\\(Tea. \& Self. Distil.)} \\
        \midrule[0.3pt]
         Acc. (\%)  & 92.6  & 93.9  & 92.6  & 94.1  \\ 
        \bottomrule[1pt]
    \end{tabular*}
    \vspace{5pt}
    \caption{Ablation study results of the distillation framework. The table presents comparative performance of four configurations.}
    \label{tab-ablation}
\end{table}

Upon reviewing the performance of Model-1 and Model-3, it is evident that Model-1, due to significant compression, exhibits limited performance compared to PointViG~\cite{2024PointViG}. The nearly identical accuracy between Model-1 and Model-3 can be attributed to the lack of guidance from a teacher model, which results in a slower convergence of the student model, thereby preventing the negative-weight self-distillation from exerting its regularizing effect on training. The comparison between Model-2 and Model-4 reveals that, with the guidance of the teacher model PointViG~\cite{2024PointViG}, Model-2 achieves performance close to that of PointViG~\cite{2024PointViG}, despite a substantial reduction in parameters. Moreover, PointViG-Distil indicates that when the student model's performance approaches saturation, the self-distillation mechanism contributes to further enhancement in performance.

The ablation study provides compelling evidence that the combined application of teacher-student and negative-weight self-distillation methods in PointViG-Distil yields the highest accuracy, surpassing the individual effects of each distillation technique. This synergistic effect underscores the significance of the distillation framework, which optimizes model performance and concurrently reduces model complexity, achieving an optimal balance between them.

\subsection{Effects of Distillation Weights on Model Performance:}
This section investigates the impact of distillation weights on the classification performance of the student model. The experimental design follows conventional settings in knowledge distillation, where the teacher-student distillation weight \( \alpha \) is set to a high value without self-distillation. This ensures that the weight is comparable to the classification loss weight, providing strong guidance for the student model. Once the teacher-student distillation parameters are established, the model further introduces a self-distillation with a negative weight \( \beta \) of low magnitude to prevent interference with the training of the student model. This section explores the effects of varying weights on model performance through a series of experiments. The results are presented in Tab.~\ref{tab-alpha} and \ref{tab-beta}, corresponding to the accuracy of the student model under different \( \alpha \) and \( \beta \) values, respectively.

\begin{table}[ht]
    \centering
    \newcommand{\tabincell}[2]{\begin{tabular}{@{}#1@{}}#2\end{tabular}}
    \small
    \setlength{\tabcolsep}{5.5mm}
    \begin{tabular*}{\textwidth}{@{\extracolsep{\fill}}c|ccccc}
        \toprule[1pt]
        $\alpha$                & 1.0   & 2.0   & 3.0   & 4.0   & 5.0   \\
        \midrule[0.3pt]
        Acc.  (\%)              & 93.7  & 93.9  & 93.7  & 93.7  & 93.5  \\
        \bottomrule[1pt]
    \end{tabular*}
    \vspace{5pt}
    \caption{Accuracy variation of the student model with different teacher-student distillation weights $\alpha$ (without self-distillation), at a constant temperature $T_\text{tea} = 3.0$.}
    \label{tab-alpha}
\end{table}

\begin{table}[ht]
    \centering
    \newcommand{\tabincell}[2]{\begin{tabular}{@{}#1@{}}#2\end{tabular}}
    \small
    \setlength{\tabcolsep}{4.5mm}
    \begin{tabular*}{\textwidth}{@{\extracolsep{\fill}}c|ccccccc}
        \toprule[1pt]
        $\beta$                 & -1.0  & -0.1  & -0.01 & 0.01  & 0.1   & 1.0   \\
        \midrule[0.3pt]
        Acc. (\%)               & 93.9  & 93.8  & 94.1  & 93.7  & 93.6  & 93.7  \\
        \bottomrule[1pt]
    \end{tabular*}
    \vspace{5pt}
    \caption{Accuracy of the student model at various self-distillation weights $\beta$, with a fixed temperature $T_\text{self}$ of 3.0. The results indicate higher accuracy with negative $\beta$ values, suggesting a regularizing effect.}
    \label{tab-beta}
\end{table}

As shown in Tab.~\ref{tab-alpha}, without self-distillation, the accuracy varies with different \( \alpha \) values, indicating that the strength of guidance has a direct impact on the knowledge transfer from the teacher model and the learning of the student model. The table demonstrates that an appropriately high \( \alpha \) is crucial for effective knowledge distillation and student model guidance.

Tab.~\ref{tab-beta} illustrates that when the teacher model provides strong guidance, the model's performance is generally better with a negative self-distillation weight \( \beta \) than with a positive \( \beta \). This confirms that negative-weight self-distillation serves a regularizing role, enhancing the model's ability to generalize from training data and preventing rapid convergence to suboptimal solutions under the teacher model's guidance.

\subsection{Visualization Analysis of Encoder Features}
\begin{figure}[htb]
    \centering
    \includegraphics[width=0.99\linewidth]{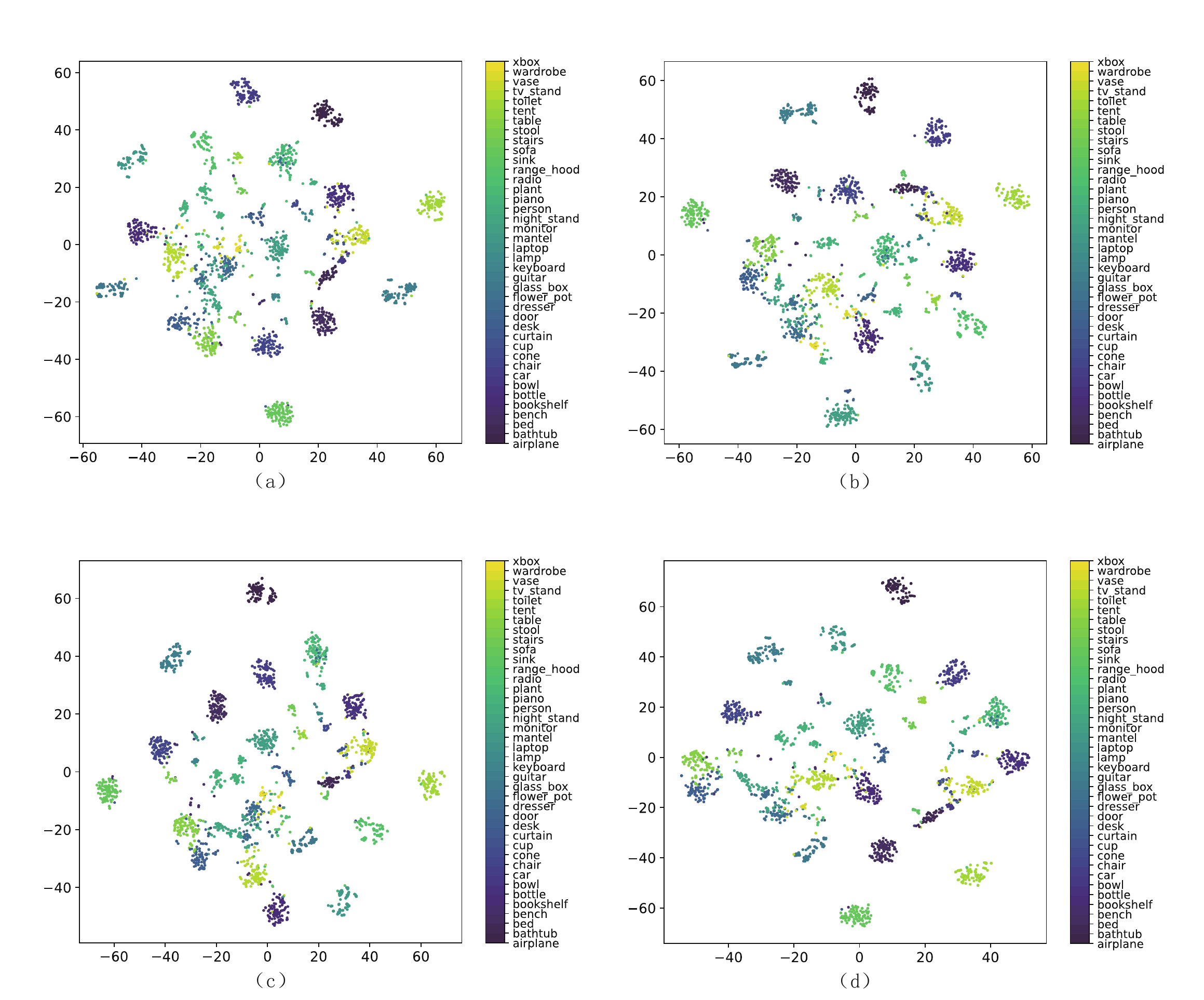}
    \caption{t-SNE visualization of encoded features for (a) PointViG~\cite{2024PointViG} (teacher model), (b) PointViG-Distil (no distillation), (c) PointViG-Distil (teacher-student distillation only), and (d) PointViG-Distil (teacher-student and negative-weight self-distillation). This figure is best viewed in an enlarged format for clarity.}
    \label{fig-PointDistil-feature-tsne}
\end{figure}

In this section, we present a t-distributed Stochastic Neighbor Embedding (t-SNE) visualization analysis of four models for point cloud classification. The models include: (a) the original teacher model, PointViG~\cite{2024PointViG}; (b) PointViG-Distil, a compressed version without any distillation techniques; (c) PointViG-Distil with only teacher-student distillation; and (d) the full PointViG-Distil model incorporating both teacher-student distillation and negative-weight self-distillation. Each model's encoded features are visualized in separate plots, labeled (a), (b), (c), and (d) in Fig.~\ref{fig-PointDistil-feature-tsne}, respectively. 

In Fig.~\ref{fig-PointDistil-feature-tsne}(a), the PointViG teacher model shows well-defined clusters, reflecting strong intra-cluster aggregation and distinct class separability, setting the benchmark for comparison. Moving to Fig.~\ref{fig-PointDistil-feature-tsne}(b), where no distillation operations are applied, the student model exhibits reduced clustering efficiency. Several clusters are less compact, and some have more stray points, indicating weaker feature representations and reduced class separability.

Fig.~\ref{fig-PointDistil-feature-tsne}(c), representing the student model with only teacher-student distillation, shows significant improvement over Model-b, closely mimicking the clustering behaviour of the teacher model. Compared to Model-b, the clusters of Model-c exhibit tighter formations and clearer separations, demonstrating that the distillation process helps the student model replicate the teacher’s feature distribution, enhancing its representational power.

Fig.~\ref{fig-PointDistil-feature-tsne}(d) depicts the full PointViG-Distil model, incorporating both teacher-student and self-distillation. Although the accuracy of this model is marginally lower than the teacher model, the t-SNE visualisation reveals more balanced cluster distributions. Compared to Model-c, the clusters of PointViG-Distil exhibit fewer outliers and clearer boundaries, reducing overlap between classes. This improvement in cluster separability and reduction in free points confirms the efficacy of negative-weight self-distillation in improving the generalization capacity of the student model.

In summary, the comparison across these four models highlights the progression in feature distribution quality with the addition of distillation techniques, validating the role of both teacher-student and negative-weight self-distillation in improving the robustness and generalization of PointViG-Distil.

\subsection{Visualization Analysis of Logit Outputs}
\begin{figure}[ht]
    \centering
    \includegraphics[width=0.99\linewidth]{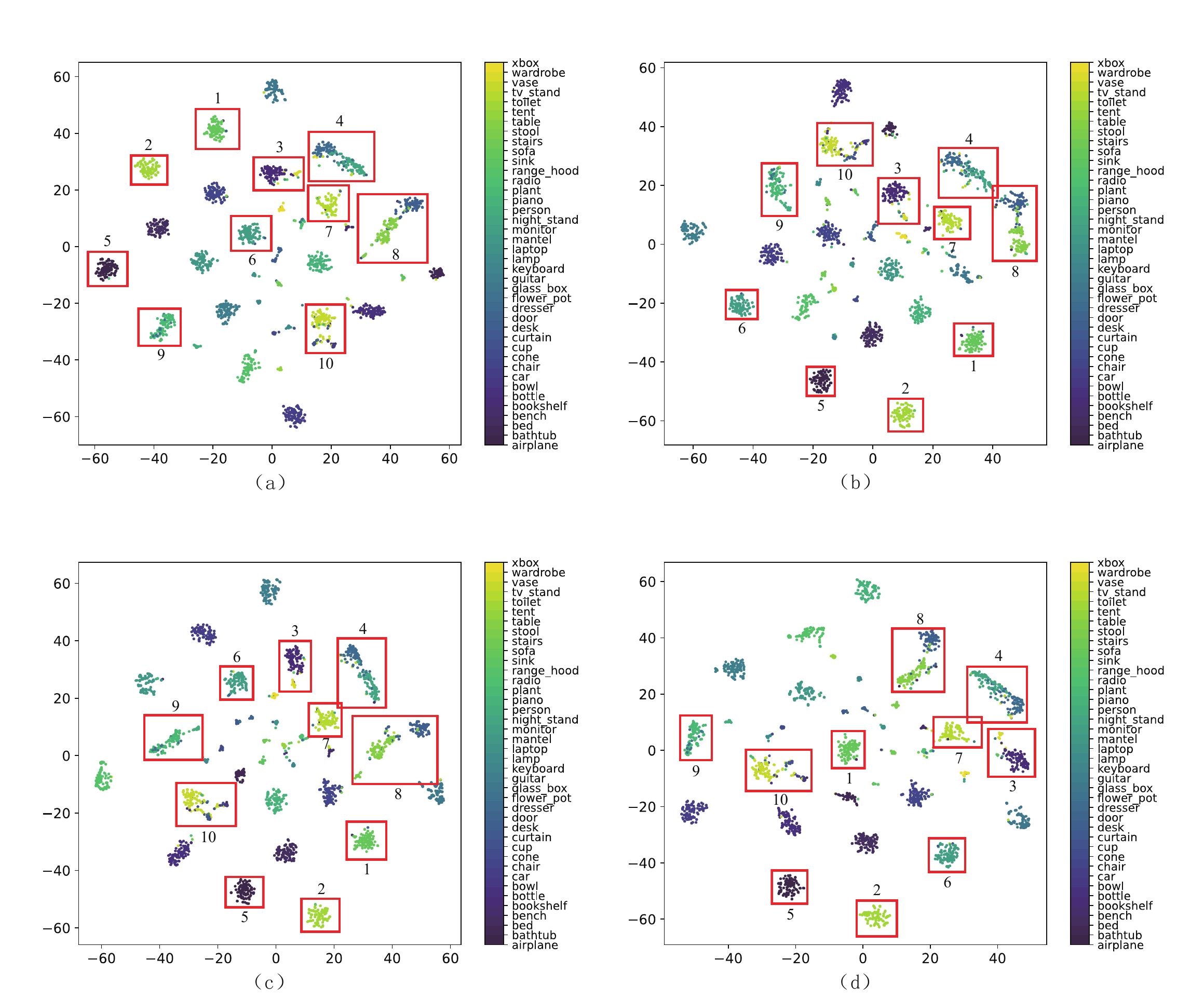}
    \caption{t-SNE visualization of encoded features for (a) PointViG~\cite{2024PointViG} (teacher model), (b) PointViG-Distil (no distillation), (c) PointViG-Distil (teacher-student distillation only), and (d) PointViG-Distil (teacher-student and negative-weight self-distillation). Ten representative regions are consistently labeled across all plots for comparison of confidence levels and decision boundaries. This figure is best viewed in an enlarged format for clarity.}
    \label{fig-PointDistil-logit-tsne}
\end{figure}

In this section, we present a t-SNE visualization analysis of the logit outputs for four models in the context of point cloud classification. The models include: (a) the original teacher model, PointViG~\cite{2024PointViG}; (b) PointViG-Distil, a compressed version without any distillation techniques; (c) PointViG-Distil with only teacher-student distillation; and (d) the full PointViG-Distil model incorporating both teacher-student distillation and negative-weight self-distillation. Each model's logits are visualized in separate plots, labeled (a), (b), (c), and (d) in Fig.~\ref{fig-PointDistil-logit-tsne}, respectively. Ten representative clusters were identified and consistently labeled across all visualizations to facilitate a direct comparison of the models' confidence levels and decision boundaries.

The visualizations revealed clear differences in the confidence and decision boundary delineation across models. \textbf{Model-a}, the teacher model PointViG~\cite{2024PointViG}, showed compact and well-delineated clusters. This tight grouping reflects high classification confidence, as the decision boundaries between classes are sharp, and the logits for each class are confidently grouped. The clear boundaries indicate that the teacher model makes strong, well-defined decisions, leaving little ambiguity between classes.

In contrast, \textbf{Model-b}, the compressed version without any distillation techniques, exhibited more dispersed clusters with fuzzier boundaries. This dispersion indicates a drop in classification confidence and less sharply defined decision boundaries. The logits from this model show increased overlap between categories, suggesting that the compressed model without distillation struggles to maintain confident, unambiguous decision-making between categories, leading to more uncertainty near the decision boundaries.

\textbf{Model-c}, which reintroduces teacher-student distillation, shows significant improvement in confidence and decision-making clarity. The clusters, particularly 1, 2, 3, 4, 6, 7, and 9, more closely resemble those from Model-a. This indicates that the addition of teacher-student distillation helps the student model mimic the teacher’s decision boundaries and confidence levels effectively. The model recaptures much of the teacher's ability to make confident and well-separated decisions, reducing overlap between different classes and enhancing decision reliability.

\textbf{Model-d}, the full PointViG-Distil model incorporating both teacher-student and self-distillation, further refines the logit space. The overall distribution of clusters is more spread out compared to Model-c. This spread indicates that the model allows for more flexibility in its decision boundaries, while maintaining high confidence in its classifications. Importantly, the overlap of points between different categories is minimized, particularly in critical areas like clusters 1, 2, 5, 7, and 10. This suggests that the model achieves a balance between strong classification confidence and broader decision regions, allowing for improved handling of difficult-to-classify samples compared to Model-c.

The t-SNE visualization analysis of the logit outputs highlight the progressive improvements from removing distillation (Model-b) to adding teacher-student distillation (Model-c) and ultimately introducing negative-weight self-distillation (Model-d), each contributing to the model's ability to make confident and precise decisions.

\section{Conclusion}
The PointViG-Distil model introduced in this study effectively enhances point cloud classification through an efficient offline distillation framework combined with negative-weight self-distillation. This approach produces a model that matches the accuracy of its teacher while significantly reducing the parameter count, marking a critical advancement for applications in resource-constrained environments. However, the model faces limitations in dense prediction tasks due to the extensive size of offline records. Future research should focus on refining these records for improved scalability. Additionally, integrating our distillation strategies into other domains, such as image classification, presents a promising avenue for further exploration. This research not only advances the field of point cloud analysis but also lays the groundwork for optimizing knowledge distillation across a variety of resource-intensive tasks, thereby enhancing model performance and efficiency in various resource-limited settings.

\bibliographystyle{elsarticle-num}
\bibliography{reference}
\end{document}